\pdfoutput=1

\documentclass[11pt]{article}

\usepackage{acl}

\usepackage{times}
\usepackage{latexsym}
\usepackage{amssymb}
\usepackage{pifont}
\usepackage{graphicx}
\usepackage{booktabs}
\newcommand{\qedb}{QEDB}

\makeatletter
\def\blfootnote{\xdef\@thefnmark{}\@footnotetext}
\makeatother
\usepackage[T1]{fontenc}

\usepackage[utf8]{inputenc}

\usepackage{microtype}

\title{QA Is the New KR: Question-Answer Pairs as Knowledge Bases}

\author{Wenhu Chen ~ William W. Cohen ~ Michiel De Jong ~ Nitish Gupta \\ \textbf{~ Alessandro Presta ~ Pat Verga  ~ John Wieting}*\\
\texttt{\normalsize \{wenhuchen,wcohen,msdejong,guptanitish,apresta,patverga,jwieting\}@google.com}\\
Google AI \\
}

\begin{document}
\maketitle
\begin{abstract}
In this position paper, we propose a new approach to generating a type of knowledge base (KB) from text, based on question generation and entity linking.  We argue that the proposed type of KB has many of the key advantages of a traditional symbolic KB: in particular, it consists of small modular components, which can be combined \emph{compositionally} to answer complex queries, including relational queries and queries involving ``multi-hop'' inferences.  However, unlike a traditional KB, this information store is well-aligned with common user information needs.
\end{abstract}

\blfootnote{*Authors listed alphabetically by last name.}
\section{Introduction: QA Pairs as a KB \label{sec:intro}}

\subsection{Background}

\begin{figure}[tb]
    \centering
    \includegraphics[width=0.4\textwidth]{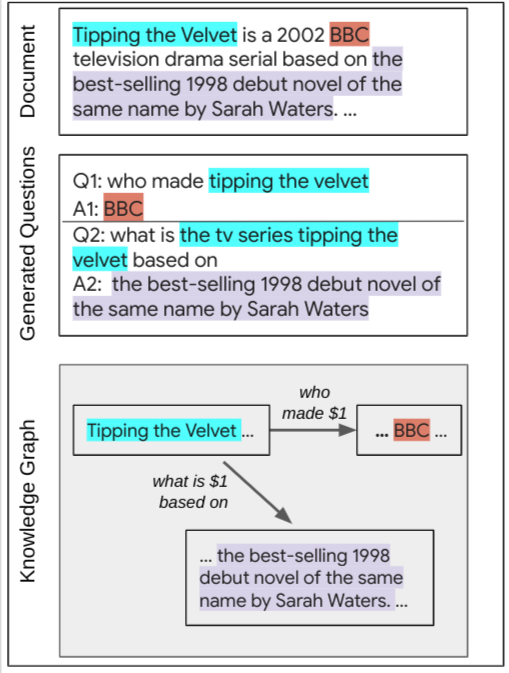}
    \caption{Constructing a \qedb{} from generated questions and explanations.  The first step is to generate question/answer pairs, and align referentially equivalent spans in the question and document (aligned spans have the same color). The aligned document spans, plus the answer spans, become nodes in a graph, and edges are labeled by an abstract versions of the corresponding questions.}
    \label{fig:small-qedb}
\end{figure}

Since the very beginnings of Artificial Intelligence (AI) research, symbolic knowledge bases (KBs) have been used for knowledge-intensive tasks.  There are many kinds of KBs, but they are based on one central principle--information is stored in small modular components (e.g., entities, KG triples, WikiData statements) that can be combined compositionally to answer complex queries. 

Broad-coverage KBs like FreeBase \cite{bollacker2008freebase} and WikiData \cite{vrandevcic2014wikidata} continue to be widely used in AI and Natural Language Processing (NLP). However, many of the facts in these KBs are of little interest to most users, and conversely, there are many facts of interest that are either not represented in the KB, or are difficult to extract from the KB without complex queries.  This is a natural consequence of how broad-coverage KBs are created:
historically, most broad-coverage KBs were populated in a ``information-driven'' way, by first identifying available information sources, and then integrating these sources such that they could be jointly queried.\footnote{This process is quite different from that used in traditional small-coverage domain-specific KB design, which generally included reflection on the needs of specific end users.}  An alternative would be a ``user-driven'' process would be to start with data on users' information needs, and use this data to drive the collection and integration of information.  In this position paper, we argue that \emph{the availability of large question-answering (QA) datasets now enables a ``user-driven'' KB development process}, and discuss the implications of this statement.

In particular, we begin with a concrete proposal for how training data for an extractive QA system---i.e., a collection of question-answer pairs, where the answers are spans of text in a corpus---can be be used to produce a KB-like structure that we call a Question-answer Explanation Database (\qedb{}).  By construction, the \qedb{} is unlikely to contain facts that are true but irrelevant to user's needs, because nobody will ask about them, and more likely to contain information that is organized similarly to user queries.  However, a \qedb{} is not simply a collection of questions and answers: we will demonstrate that elements of \qedb{} can be combined compositionally, just as knowledge graph (KG) triples can be combined, and we argue that this leads to a succinct and natural query language.  We argue that such a KB is better-matched to users' information needs, and support this claim with experimental evidence.

\subsection{Question-answer explanations as a KB}

Our approach to generating a KB from raw text exploits recent progress in question generation (QG) \cite{yang2017semi,lewis2021paq} and explainable question-answering \cite{lamm2021qed}.  The approach is illustrated in Figure~\ref{fig:small-qedb}, and consists of the following steps.

\begin{figure*}[tb]
    \centering
    \includegraphics[width=1.0\textwidth]{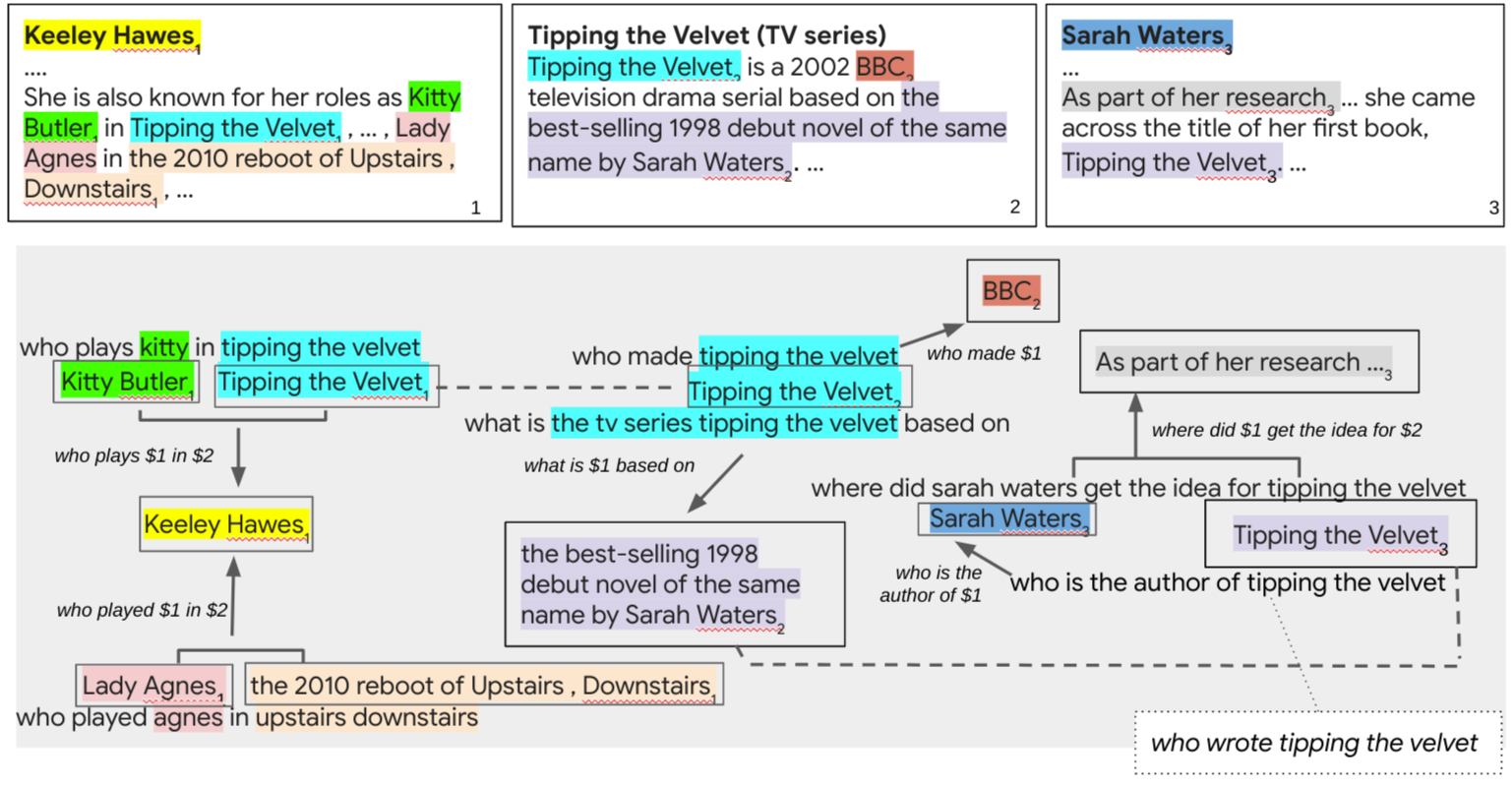}
    \caption{A \qedb{} derived from three documents.  Generated questions are placed near the document spans aligned with them, and the subscript on a document span indicates its containing document. Dashed lines (and shared colors) indicate cross-document co-reference, and dotted lines indicate semantically equivalent questions.  Note that the extractions from documents 1 and 3 include ternary relations.}
    \label{fig:large-qedb}
\end{figure*}


(1) Given a document $d$, a large number of question-answer pairs $(q,a)$ are generated, where $a$ is a span from $d$.  In Figure~\ref{fig:small-qedb}, the questions were generated automatically from a sequence-to-sequence neural model trained on a subset of the Natural Questions (NQ) dataset \cite{kwiatkowski2019natural}.

(2) For each question-answer pair $(q,a)$, we align certain referential spans $r_q$ in the question with \textit{referentially equivalent} spans $r_d$ in the document. For instance, in Figure~\ref{fig:small-qedb}, the two light blue spans ``\textit{the tv series tipping the velvet}'' in $q$ and ``Tipping the Velvet'' in $d$ are aligned, because their referents in the real world are equivalent.  We call this alignment a \textit{question explanation}, as it partially justifies why the answer $a$ is supported by the document $d$ \cite{lamm2021qed}.

(3) Finally, a graph is formed by constructing a node for each answer span $a$ and each question-aligned reference span $r_d$ in the document. Nodes associated with the same question $q$ are then connected by an edge that is labeled with a ``relation'' associated with $q$'s surface form. For instance, the question ``\textit{what is the tv series tipping the velvet based on}'' is used to produce the edge label $q_l=$``what is \$1 based on''. These $(r_d, q_l, a)$ triples are analogous to a $(subject, relation, object)$ found in a typical KB. 
As shown in Figure~\ref{fig:small-qedb}, these edges typically indicate salient relationships from the document.

(4) A richer graph structure can be produced by running an entity linker \cite{roth2014wikification} on the underlying corpus of passages, since many question-aligned spans correspond to or contain linkable entities.\footnote{In our data, almost 3/4 of the questions contain references that can be linked, and more than half contain both an answer and question reference that can be linked.  We allow links between a reference and an entity properly contained in that reference (e.g., between ``\textit{the race in grease}'' and the movie \textit{Grease}) but such proper containments are uncommon in the data, with almost 90\% of a sample of linkable references having high similarity to the linked entity.}  
Figure~\ref{fig:large-qedb} shows a larger \qedb{} derived from information in three documents, with dashed lines indicating passage references mentioning a common entity. (For compactness, the generated questions are shown next to the document nodes to which they align.)

The resulting collection of question-answer pairs, together with their entity annotations, is the \qedb{}.  More formally the atomic elements of the \qedb{} each contain: a question $q$; an answer $a$, which is a span in a document $d$; two lists of referential spans $r_q^1, \ldots, r_q^k$ and $r_d^1, \ldots, r_d^k$, where each $r_q^i$ is a span in $q$, each $r_d^i$ is a span in $d$, and $r_q^i$ and $r_d^i$ have the same referent; and a list of sets of entities $E^1, \ldots, E^k$, where $E^i$ is the set of entities mentioned in the span $r_d^i$.

The \qedb{} is strongly influenced by the initial sample of questions; grounded in the corpus; relational, containing an open set of relations; and (since it contains traditional KB entities) it can be easily combined with a symbolic KB over the same entities.

Notice that the relational graph structure arises in two ways: because of the entity linking, and because multiple generated questions can be aligned with the same document span.  The graph structure makes it possible to use the \qedb{} to answer questions more complex than the original generated questions: for example, a question like ``where can i watch that tv series based on a 1998 novel by sarah waters'' could potentially be answered by the \qedb{} of Figure~\ref{fig:small-qedb}.

Figure~\ref{fig:large-qedb} also shows that the \qedb{} formalism naturally models relationships with three or more arguments, since questions can have multiple reference spans---e.g., the question ``\textit{who plays kitty in tipping the velvet}'' has two reference spans, aligning to ``\textit{Kitty Butler$_1$}'' and ``\textit{Tipping the Velvet$_1$}'').  In the figure, non-binary relationships are shown as directed hyperedges, connecting two question-aligned spans to an answer span.


\subsection{Discussion: \qedb{}'s versus open IE}

The process shown in Figure~\ref{fig:small-qedb} is closely related to information extraction (IE) techniques \cite{sarawagi2008information}---it is especially similar to open IE methods \cite{etzioni2008open,mausam2016open}, since there is no fixed vocabulary of entities or relations.  However, it differs in being driven by an initial QG stage.  If we assume that QG produces questions representative of real users' questions, then this initial stage serves an important role: \emph{it ensures the information in the KB corresponds to realistic information needs} of the users.  This is in contrast to existing broad-coverage KGs, which we claim are \emph{not} well-matched to user's needs, and with OpenIE triples, which are often not informative. Below we will provide additional evidence for this claim, but consider for now one concrete example.  The first question from the NQ development set is:
\textit{who got the first nobel prize in physics?} For most reasonable KB schemata, this would lead to a complex KB query.\footnote{Specifically, consider the intricacies of querying a potentially incomplete KB to find the ``first'' item in a series.}  In contrast, a \qedb{} based on this question and document that contains the answer (i.e., ``\textit{The first Nobel Prize in Physics was awarded in 1901 to Wilhelm Conrad Röntgen \ldots}'')  would contain the edge

\begin{center}
\begin{tabular}{c}
\framebox{The first Nobel Prize in Physics} \\
$\downarrow$~~{\scriptsize who got \$1} \\
\framebox{Wilhelm Conrad Röntgen}
\end{tabular}
\end{center}
 
To summarize, the \qedb{} can potentially combine key advantages of symbolic KBs and open QA systems \cite{lee2019latent,karpukhin2020dense,mao2020generation,izacard2020leveraging}: like open QA methods, they can exploit the fact that unstructured text is often well-aligned with common questions, and like symbolic KBs, they make it possible to efficiently answer complex queries, by navigating through extracted information, or aggregating information from different parts of a KB.

The issue of how well-aligned \qedb{} information is to user questions will be revisited with experiments in Section~\ref{sec:qedb-utility} below.

\section{Methods}

Although primarily a position paper, some of the claims we make here can be supported experimentally.
The experiments of this paper are conducted with a particular \qedb{}.  Below we outline how it was constructed, and the computational methods used.

\subsection{Question generation and filtering}

The questions in the \qedb{} are taken from the Probably-Asked Questions (PAQ) dataset introduced by \citet{lewis2021paq}.  PAQ contains 64.9 million questions generated from Wikipedia passages (specifically, the widely-used 20 million passages from \citet{karpukhin2020dense}). The dataset was created via a four-stage pipeline. (1) In \textit{passage selection}, a model scores each passage by how likely it is to contain useful questions. This model is trained as a binary-classifier using the answer-containing passages from NQ as positives and heuristically selected other passages as negatives. (2) In \textit{answer extraction}, a second model takes the top $n$ scoring passages from stage one and predicts likely answers, using a combination of an off-the-shelf named-entity-recognition model and an answer span selector trained using NQ short answers.  (3) In \textit{question generation}, a third model takes a passage and demarcated answer, and generates a question that should be answerable by the text. This generative model is trained by inverting standard question answer datasets - rather than producing $a=f(q, p)$, the model instead produces $q=f(a, p)$. Concretely, this is a BART model trained on a combination of SQUAD, TriviaQA, and NQ.  (4) Finally  the questions are \textit{filtered} in two ways.  First ``round-trip'' filtering is used to ensure that a standard question answering model, when given the generated question, produces the same answer that the question generator was conditioned on.  \citet{lewis2021paq} used an openQA model as the filter which produces $a=f(q, C)$ where $C$ is the entire corpus rather than the single ``gold'' passage.  This is a very selective filter, which discards more than 3/4 of the generated questions.  For more details, consult \cite{lewis2021paq}.

\subsection{Question reference to passage reference alignment and linking}

\begin{table}[!thb]
\centering
\small
\begin{tabular}{c|cccc}
\toprule
                  & \multicolumn{2}{c}{Passage Reference}   & \multicolumn{2}{c}{Question Reference} \\
                  & EM & F1   & EM & F1 \\
\midrule
T5-XXL                &  67.7  & 64.9 & 75.7 & 82.2 \\
T5-Large              &  56.5  & 54.5 & 72.7 & 79.7 \\
\bottomrule
\end{tabular}
\caption{Results on QED dev set for identifying passage and question references.}
\label{tab:qedextraction}
\end{table}

To align the question references with passage references, we trained a sequence-to-sequence model that takes a question, passage, and demarcated answer, and predicts the question references, passage references, and predicate form as a single text-to-text output. The model is a T5 model initialized from the standard checkpoints and fine-tuned using annotations for a subset of NQ \citep{lamm2021qed}, which were originally used for explainable question-answering.  The performance of this model is shown in Table~\ref{tab:qedextraction}.

An off-the-shelf entity linker was run on all the passages.  Preliminary experiments showed that linking entities in the questions directly was very inaccurate---this is unsurprising as short snippets of text like questions have limited contextual information, and context is known to be required for many cases of entity linking.  The entity-linked passage references were then matched heuristically to question references, as follows: an entity $e$ was associated with a question reference $r_q$  if (1) $e$ was linked to the passage $p$ from which the question was generated and (2) no other entity $e'$ linked from $p$ is more similar to $r_q$ than $e$.  In measuring surface similarity, we used Jaccard distance on tokens, and considered both the surface form of $e$ in $p$ and a single canonical entity name for $e$.  This simple heuristic for matching surface forms could no doubt be improved, but performs quite well, as the qualitative analysis below illustrates.

\subsection{Question-answering using the \qedb{}}

One of the ways to evaluate a \qedb{} is to simply use it as a resource for answering simple one-hop questions.  Following \citet{lewis2021paq}, we implemented a method that answers user questions $q'$ by retrieving similar question-answer pairs $(q_1,a_1),\ldots,(q_k,a_k)$, concatenating them with the original question $q'$, and then using a Transformer to fuse this information and generate a final combined answer $a'$.  The model we used, called QAMAT (for Question-Answer Memory Augmented Transformer), is described in detail elsewhere \cite{chen2022augmenting}.

QAMAT is closely related to RePAQ, the learning model of \citet{lewis2021paq}, in that it learns to neurally retrieve similar questions end-to-end as a subgoal of the QA task.  However, QAMAT also includes a novel pre-training phase, which learns to unmask salient spans in a passage using a small batch-sized memory of questions generated from that passage (and distractors).  This substantially improves question retrieval on smaller QA datasets, such as the WebQuestion variants we consider below.

\subsection{Question-answering using a KB}

For comparison purposes, we also implemented a variation of QAMAT which replaces the memory question-answer pairs with a memory of Wikidata facts, which we will call FAMAT.  Pretraining QAMAT/FAMAT requires a corpus of documents that have been heuristically aligned with the items to be retrieved, and for FAMAT we adopted the pre-training corpus used for FILM, the Fact Injected Language Model \cite{verga2021adaptable}.
Besides FAMAT, we also consider a strong baselines for QA using a traditional KB, FILM \cite{verga2021adaptable}.

\section{Experiments and Observations\label{sec:experiments}}

\begin{table*}[!thb]
\centering
\scriptsize
\begin{tabular}{cll}
Query Entity  & Related Entity & Linking Question \\
\hline
Tonga   & Deborah Ann Gardner   & \textit{in which country was deborah gardner murdered by the peace corps in 1976}\\
        & 1976  & \textit{in which country was deborah gardner murdered by the peace corps in 1976}\\
        & Fiji  & \textit{where did the sulu come from in fiji}\\
        & American Samoa        & \textit{what country did american samoa defeat to win their first ever world cup}\\
        & 2011  & \textit{what country did american samoa lose to in 2011}\\
        & New Zealand   & \textit{in 2006, australian troops and new zealand police were sent to which island}\\
        & Tropical cyclone      & \textit{which island nation was hardest hit by cyclone gita}\\
       & 1969  & \textit{which country issued a banana shaped postage stamp from 1969 to 1985}\\
        & Samoa & \textit{who did samoa play in the under 20 world cup}\\
        & Niuatoputapu  & \textit{niuatoputapu is the highest point in which pacific island nation}\\
\hline
arXiv   & Preprint      & \textit{where can you find a preprint repository}\\
        & Ricci flow and the Poinc (GPC)        & \textit{on which computer system is ricci flow and the poincaré conjecture}\\
        & An Exceptionally Simple \ldots       & \textit{where is an exceptionally simple theory of everything located}\\
        & VAN method    & \textit{where is the van method of measuring earthquakes stored}\\
        & ARC fusion reactor    & \textit{where did the idea of the arc fusion reactor come from}\\
        & Database      & \textit{what is an example of a preprint database}\\
        & Mathematics   & \textit{who invented the first open access journal in the history of mathematics}\\
        & Particle      & \textit{where was the article about quarkonium published}\\
        & Results in Mathematics        & \textit{on what platform have some important results in mathematics been published}\\
        & Particle physics      & \textit{where can preprints be found in physics}\\
\hline
Jerry Garcia    & Grateful Dead & \textit{which member of the grateful dead died in 1995}\\
        & Guitar        & \textit{who played guitar on three tracks by ornette coleman in 1988}\\
        & Lead guitar   & \textit{who was the lead guitarist for the dead in europe in 1972}\\
        & Pedal steel guitar    & \textit{who plays guitar on teach your children by graham nash}\\
        & Today (Song) & \textit{who plays the lead guitar on today by jefferson airplane}\\
        & *\underline{Roseanne} (TV show character)     & \textit{what's the name of roseanne's fourth child}
                                            *\underline{answer is actually Jerry Garcia Conner}\\
        & Teach Your Children   & \textit{who plays guitar on teach your children by graham nash}\\
        & 1995  & \textit{which member of the grateful dead died in 1995}\\
        & The Grateful Dead Movie       & \textit{who directed the movie the grateful dead movie}\\
        & Sandy Rothman & \textit{sandy rothman was a close friend and collaborator of which grateful dead guitarist}\\
\hline
Pittsburgh      & Pennsylvania      & \textit{where is kraft heinz located in pennsylvania}\\
        & United States & \textit{where was the first legion of the united states raised}\\
        & Pittsburgh Steelers   & \textit{where did the steelers play in the 1964 super bowl}\\
        & University of Pittsburgh      & \textit{where does the university of pittsburgh bus service}\\
        & Fort Duquesne       & \textit{which american city was originally called fort duquesne after it was captured by the}\\
        & Monongahela River     & \textit{which american city was founded on the river monongahela in 1837}\\
        & Allegheny County      & \textit{what is the largest city in allegheny county pennsylvania}\\
        & Ohio River    & \textit{where is the rowing race on the ohio river}\\
        & Andy Warhol   & \textit{where did andy warhol live before they moved to pennsylvania}\\
        & The Andy Warhol Museum     & \textit{where was the warhol museum in which the movie basquiat was made}\\
\hline
NC University System & Kuklos Adelphon       & \textit{where did kappa alpha and kuklos adelphon meet}\\
        & +Geoff Crompton        & \textit{where did geoff crompton play college basketball} +\underline{UNC/CH not the system}\\
        & Margaret Spellings    & \textit{margaret spellsings is the current president of which university}\\
        & President     & \textit{margaret spellsings is the president of which university}\\
        & +Bones McKinney        & \textit{where did bones mckinney play in college} +\underline{UNC/CH not the system}\\
        & North Carolina Central U      & \textit{what system did nccu join}\\
        & Winston-Salem State Univ      & \textit{winston salem state university is a constituent institution of what university}\\
        & UNC/Greensboro      & \textit{what system is the university of north carolina at greenboro part of} \\
        & North Carolina State Uni      & \textit{what is nc state part of}\\
        & East Carolina University      & \textit{who approved the east carolina university dental school}\\
        \hline
\end{tabular}

\caption{Pairs of entities considered ``related'' by the \qedb{}, and the questions that gave rise to those relationships. These pairs are obtained by finding the top $k$ questions touching a query entity, and then finding all entities touching one of these questions. We discard related entities that are years.}  \label{tab:mid2mid}
\end{table*}

The central claims made in this position paper are the following: 
\begin{itemize}
\item (P1) Like a traditional symbolic KB, \emph{a \qedb{} is composed of modular components that can be combined in compositionally in many different ways.}
\item (P2) \emph{The information in a \qedb{} will be more useful for answering user information needs} than the information in a traditionally constructed broad-coverage KB.
\end{itemize}

If these positions are generally true, this would suggest a number of new directions and emphases for research. While general statements of this sort are difficult to support conclusively, in this section we will present some qualitative and quantitative results that we believe to support this claim.

\subsection{What do the atomic elements of the \qedb{} look like?}

Like a KB, a \qedb{} also encodes relationships.  Relationships are in general between several text spans---the question references and answer for a generated question---but since many of these references can be linked to entities, the \qedb{} encodes entity-to-entity relationships as well.

To visualize the types and quality of relationships in \qedb{}, we picked five $e$ entities at random (with a constraint that one author was at least slightly familiar with them).  For each entity $e$ with found the top ten entities $e'$, ranked by the number of questions $q_{e,e'}$ that have an answer linked to $e$ and a question reference linked to $e'$ (constraining the confidence of the linking to be more than 0.25).  To summarize the relationships expressed, we list in Table~\ref{tab:mid2mid} the query entity $e$, up to ten related entities $e'$, and one associated question $q_{e',e}$. Although we limit the sample to 10 lines per query entity $e$, there is no ``cherry picking'' in this sample.

The relationships are fairly diverse and, from inspection, seem to be fairly accurate.  Of the 100 selected relationships, a few, mostly those involving years, are uninformative when shown as binary relations---it is nonsensical to say that Tonga is related to the year 1976. However, it is certainly sensible to say that Tonga is related to a newsworthy event that occurred in this year, which is what is actually encoded in the \qedb{}.\footnote{In the table we do not necessarily include all entities associated with question references, since we limit the total number of $e'$ to 10 for each query entity.}
One of the relations (marked with an asterisk) is clearly wrong, the result of an incorrect entity link between a TV show character (Jerry Garcia Conner) and a musician with a similar name (Jerry Garcia).  There are also two more subtle entity linkage errors where the entity for the University of North Carolina \emph{system}, a multi-campus institution, is conflated with the entity for its flagship campus at Chapel Hill.  These are marked with a plus in the table (while it is not technically incorrect to say that Bones McKinney played for the UNC college system, it seems at least unnatural.)

\subsection{Can elements of the \qedb{} be composed?}

One key aspect of a KB is that it contains small, modular components that can be recombined to answer complex inferences.  To explore this characteristic of a \qedb{}, we qualitatively evaluated the results of certain kinds of ways of combining components.



\begin{table*}[tb]
\centering
\scriptsize
\begin{tabular}{llll}
Question 1 & Bridge Entity & Question 2 & Answer \\
\hline
\textit{where did the cincinnati reds last game} & Joe's North ... & \textit{what nfl team used to play at \$1} & Cincinnati Bengals \\
\textit{who sings the song please leave the grates} & Jebediah & \textit{when was \$1 formed and by whom} & 1994 \\
\textit{what event was hosted in 2012 by marseille} & World Water Forum & \textit{who organizes \$1} & World Water Council \\
\textit{who is the main character of the tombs of atuan} & Tenar & \textit{who raised \$1 in wizard of earthsea lore} & Aihal \\
\textit{what is the sequel to wild fire by nelson de mille}* & Night Fall & \textit{when did the plane crash in \$1} & 1996 \\
\textit{who directed the opening act} & Steve Byrne & \textit{\$1 is the lead actor in which us tv series} & Sullivan \& Son \\
\textit{who designed the set for le piege de meduse} & Willem de Kooning& \textit{\$1 is an example of what} & abstract expressionism \\
\textit{who maintains the lights on the isle of man} & N. Lighthouse Boar & \textit{\$1 is in which country} & Scotland \\
\textit{who was the roman proponent of hedonism} & Lucretius& \textit{what is the name of \$1's book on atomism} & On the Nature of Things \\
\textit{what is the main export of tutuila} & Canned fish & \textit{what is ... the process used to preserve \$1} & Canning \\
\textit{who replaced randy jackson as mentor for this season} & Scott Borchetta & \textit{who did \$1 try to recruit to his record label} & Taylor Swift \\
\textit{where did the bengals play in the american football league} & Joe's North ... & \textit{where did the cinc. bengals play before \$1} & Riverfront Stadium \\
\textit{where is laurence harbor located in new jersey} & Raritan Bay & \textit{what do you do in \$1} & recreational fishing \\
\textit{munawar pass is famous for being the site of which operation} & Operation Gibraltar & \textit{\$1 took place in which state} & Kashmir \\
\textit{which novel features the village of puttenham, surrey}* & Brave N. World & \textit{how did george orwell describe \$1} & negative utopia \\
\hline
\textit{who was the creator of  "the rocky horror show"} & Richard O'Brien & \textit{when was \$1 born} & 1942 \\
\textit{which nickelodeon show was bill long credited in} & Blues Clues & \textit{who originally hosted \$1} & Steve Burns\\
\textit{what college was thomas balaton educated at after eton} & New College & \textit{when was \$1 founded } & 1379 \\
\textit{which actor composed the song "smile" from 1st round} & Charlie Chaplin& \textit{what nationality is \$1} & English\\
\textit{what casino is formerly known as vegas world} & Stratosphere Las Vegas & \textit{what company is headquartered at \$1} & Amer. Casino \& Entertain\ldots\\
\hline
\end{tabular}
~\\
~\\
$*$Questions marked with an asterisk have potentially erroneous bridging entities (e.g., linking the novel ``Brave New World'' with a game of the same name).\\

\caption{Above the line, the result of joining question pairs, where the answer of question 1 is a question reference in question 2. Below the line, five bridge questions from HotpotQA, manually converted to the same format.} \label{tab:hops}
\end{table*}

The primitive components stored in the \qedb{} are questions annotated with entity references.  From these, we constructed more complex queries by combining together pairs of questions $q_1$, $q_2$ where the answer to $q_1$ appears as a question reference in $q_2$.  For example for $q_1$=``\textit{who was the roman proponent of hedonism}'' the answer is \textit{Lucretius}, which appears in $q_2$=``\textit{what is the name of lucretius's book on atomism}''.  Joining these together yields a query like a HotPotQA ``bridge question'' \cite{yang2018hotpotqa}: expressed in natural language, the combined query might be written \textit{what is the name of the book on atomism written by a roman proponent of hedonism}.

Many hundreds of millions of plausible bridge questions can be constructed this way.  We present a sample of them in Table~\ref{tab:hops}, with the bridging entity's mention in $q_2$ replaced with a variable \textit{\$1}.  This formalism loosely follows the question decomposition meaning representation (QDMR) language proposed in \citet{wolfson2020break}.  (The lack of case, and the occasional disfluencies in the question, is similar to the NQ questions used to train the question generator for \qedb{}.) 

Specifically, the sample was selected from question pairs where (1) $q_2$ contains a single question reference; (2) each question has a single answer; (3) the confidence of the question-reference to passage reference alignment is at least $\frac{2}{3}$;  (4) the answer to $q_1$ (the bridge entity) is not a year, and is relatively uncommon (as a bridging entity, it cannot appear more than about 100,000 times in the full set of question joins); and (5) the answer to $q_2$ does not appear in $q_1$.  The first two of these constraints were imposed for convenience in processing, the third to avoid low-confidence alignments, and the fourth to promote diversity in the examples, as there are quadratically many\footnote{If an entity appears $n$ times as an answer and $n$ times as a question reference then there will be $n^2$ possible 2-hop questions.} question pairs bridged by popular entities. (E.g., ``\textit{who wrote blowing in the wind}'' and ``\textit{what was bob dylan's second album called}'').\footnote{Even with the constraint, one bridge entity appears twice in the sample.}

A total of 20 questions were selected randomly, of which five were discarded because they were too long to fit easily in the table.  For comparison, we also include five bridging questions from HotpotQA, manually converted to a similar format.  The \qedb{}-produced queries appear to be similar in quality to the HotpotQA queries, hence the \qedb{} also \emph{implicitly contains answers to many plausible multi-hop information needs}, just as a symbolic KB does.

Even with all the constraints employed above (many of which could be easily relaxed), there are still more than 6 times as many \qedb{}-generated bridging questions as there are in HotpotQA.  Of course, unlike the HotpotQA queries, these queries are expressed only in the semi-structured format shown in the table, not in purely textual form (e.g., the first HotpotQA question is actually ``\textit{when was the creator of the rocky horror picture show born}'').  


\begin{table}[h]
    \centering \small
    \begin{tabular}{cc}
         &  $P(a'=a | q'=\textit{top}_1(q))$ \\
         \hline
     BM25    & 23.9 \\
     QAMAT retriever & 40.4 \\
     RePAQ retriever & 41.6 \\
     MGT similarity & 41.8 \\
     \hline
    \end{tabular}
    \caption{Accuracy of question-similarity methods.  Following \citet{lewis2021paq}, we use NQ test questions $q$ as probes, and measure the probability that the most similar question $q'$ in
    the \qedb{} has the same answer as $q$.
    }
    \label{tab:retrieve}
\end{table}

Table~\ref{tab:hops} only shows 2-hop bridging questions, which is one particular way of joining questions pairs---i.e., we find $q_1$, $q_2$ where the answer to $q_1$ appears as a question reference in $q_2$.   There are several other ways that two questions could share an entity:
\begin{itemize}
    \item An entity $e$ could appear as a \emph{question reference} in two different questions $q_1$, $q_2$, for example ``\textit{what nfl team used to play at $e$}'' and ``\textit{what baseball team played their last game at $e$}''.  This is similar to the way information is organized in a WikiData frame for $e$, with questions serving the role of relations, and the answers acting as slot fillers.

    \item An entity $e$ could be an \emph{answer} to two different questions $q_1$, $q_2$, for example $q_1$ = ``\textit{where did the cincinnati reds play their last game}'' and $q_2$ = ``\textit{where did the bengals play in the american football league.}''  This pattern corresponds to a ``Type II'' question in HotpotQA.
\end{itemize}

We will not present examples of these sorts of questions, but they would be straightforward to produce: the only additional component needed is a test to see if two questions are ``different''.  This could be done by using the similarity metric used in the question retriever latently learned by QAMAT (or the similar retriever learned latently by RePAQ), or off-the-shelf sentence similarity metrics such as \cite{wieting2019bilingual}.  We present some quantitative results on question retrieval in Table~\ref{tab:retrieve}. (Results on RePAQ are from \cite{lewis2021paq}; MGT is a scaled-up version of the method of \citet{wieting2019bilingual}.)  We leave further investigation of these types of complex questions to further work.

Taken together, these results support claim P1: Like a traditional symbolic KB, a QEDB is composed of modular components that can be combined compositionally in many different ways.  In particular, the individual relationships in the QEDB are meaningful and correct (in 47/50 hand-checked examples).  Further, relationships can chained together to construct paths through the QEDB that are semantically similar to HotpotQA multi-hop questions.

\begin{table*}[tb]
    \centering
    \small
    \begin{tabular}{lrrrrrr}
	                 \multicolumn{3}{r}{WebQSP/WD~~~~~}  
	                & \multicolumn{2}{c}{WebQSP} 
	                & \multicolumn{2}{c}{TriviaQA} \\
	                 \multicolumn{3}{r}{\it (ideal for KBQA)}  
	                & \multicolumn{2}{c}{\it (good)} 
	                & \multicolumn{2}{c}{\it (neutral)} \\
	           \hline
	        & \multicolumn{6}{c}{\it models using a \qedb} \\
    QAMAT (ours) 	& 61 &	& 56 &	& \textbf{53} \\
    \hline
    	    & \multicolumn{6}{c}{\it models using a traditional KB} \\
    FILM     & \textbf{80} &(+19)	& \textbf{57} & (+1)	& 29 & (-24) \\
    FAMAT (ours)    & 59 &(-2)  & 50 & (-6)   & &\\
    \hline
    \end{tabular}~~~~~~\raisebox{-0.5\height}{\includegraphics[width=0.4\textwidth]{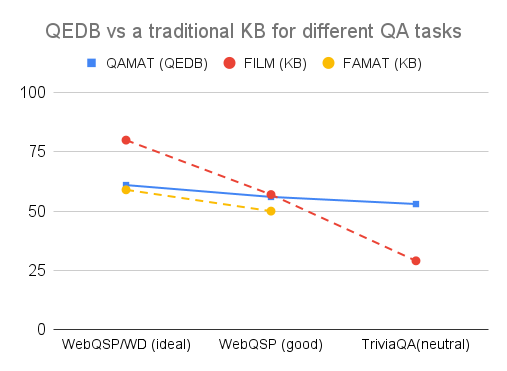}}
    \caption{Exact match performance of QAMAT, a system using a \qedb{} as a KB, against two strong baselines designed to answer questions using a traditional symbolic KB.  In the figure on the right, dashed lines are systems using a KB, and the solid line for QAMAT, which uses a \qedb{}.  The blue lines are associated with QAMAT and its KBQA variant FAMAT.}
    \label{tab:webq-vs-tqa}
\end{table*}

\subsection{How useful are QA pairs compared to KB triples?} \label{sec:qedb-utility}

To argue for P1, we evaluate performance on three question-answering datasets, which vary in how well-aligned one would expect them to be to a symbolic KB.

\textbf{TriviaQA} \cite{joshi2017triviaqa} is a widely-used question-answering dataset composed of questions authored by trivia enthusiasts.  We select this as a QA dataset that is plausibly reflective of user information needs.  While other QA datasets, notably NQ, are arguably more representative of common user questions, we feel the choice of TriviaQA is much more favorable to KBQA approaches than NQ: in particular, around 85\% of the answers in TriviaQA are KB entities \cite{fevry2020entities}, while only about 40\% of the answers in NQ are.

\textbf{WebQSP} is \cite{yih2016value} is the Web Questions Semantic Parse dataset.  We select this as a benchmark that is not reflective of user needs, but designed to be favorable to QA methods that use traditional broad-coverage KBs.  WebQSP is a subset of the WebQuestions dataset \cite{berant2013semantic}, which is also very favorable to traditional KBs, in several ways.
\begin{itemize}
    \item Questions were selected to be entity-oriented: in particular, they were produced using the Google Suggest API and constrained to begin with a wh-word and contain exactly one entity, and they were produced using a heuristic crawl starting with the question ``When was Barack Obama born?"
    \item Of these selected questions, crowd workers were explicitly instructed to find answers using the FreeBase page of the question entity, and discard questions that were not answerable this way.  This process (together with a check on annotator agreement) reduced the initial set of 100,000 Google Suggest questions by more than 90\%.
\end{itemize}
WebQuestionsSP additionally has the property that semantic parses (as SPARQL queries) could be found for the question.

Finally, \textbf{WebQSP/WD} selects out questions from WebQSP which can be aligned\footnote{In the sense that there is a WikiData statement linking the question entity and answer entity.} with a subset of WikiData statements involving 1 million ``head'' WikiData entities. Note that WebQSP is based on FreeBase, an older KB, and WikiData \cite{vrandevcic2014wikidata} is a much newer broad-coverage KB, one which is arguably better-designed and more useful than FreeBase. Also (because it is more recent) is easier to integrate WikiData with modern NLP tools like language models.  WebQSP/WD, together with the accompanying KB of WikiData facts, is selected as an ideal case for QA using a KB: carefully selected KB-oriented questions, accompanied by a tractable subset of a modern symbolic KB.

Table~\ref{tab:webq-vs-tqa} presents results on each of these datasets for QAMAT, which uses the \qedb{}, and FILM, which uses a KB. 
On WebQSP/WD, which we take to be the ideal case for QA using a KB, FILM performs much better than QAMAT.  However, FILM performs quite similarly for the full WebQSP, which contains very KB-oriented questions posed against an imperfect KB, and  FILM performs much worse on TriviaQA, trailing QAMAT by more than 20 points.

Of course, FILM and QAMAT are very different systems: for instance, FILM outputs an entity identifier an answers, rather than generating text.  The last line of the table presents performance for the same model used in QAMAT, but populated with a memory of facts instead of question-answer pairs (FAMAT).  In this comparison, \emph{where only the information store is varied}, we see that FAMAT is actually slightly worse than QAMAT, even on the ``ideal'' and ``good'' cases of WebQ\-SP/WD and WebQ\-SP.

The results of this section thus support claim P2: The information in a QEDB is indeed more useful for answering typical user information needs than the information in a traditionally constructed broad-coverage KB, as witnessed by the 24 point improvement on TriviaQA.  In fact, even on collections of questions like WebQSP that have been aggressively filtered to be KB-relevant, the QEDB is competitive with the traditional KB---only when one ensures a nearly perfect alignment between questions and KB (WebQSP/WD) does the traditional KB perform better.

\section{Related Work \label{sec:related_work}}

\subsection{Extracting and querying structured KBs} 

Beginning in the 1990s large numbers of structured databases began to be hosted on the web, largely as semi-structured text, leading to extensive research on data integration \cite{hearst1998information}.  Complemented by advances in information extraction from unstructured text \cite{sarawagi2008information}, this work eventually led to construction of large-scale, broad-coverage KGs such as DBPedia \cite{auer2007dbpedia}, FreeBase \cite{bollacker2008freebase}, 
YAGO \cite{rebele2016yago}, NELL \cite{mitchell2018never}, and WikiData \cite{vrandevcic2014wikidata}.  While these projects had different emphases and employed different mixes of technology, all of them made extensive use of semi-structured sources for data\footnote{Even NELL, which was perhaps most focused on IE from free text, made heavy use of techniques for IE from semi-structure text \cite{wang2008iterative}.}.  While intuitions about what information would be most useful for end users certainly affected the design of these KGs, little data about user needs existed at this time, and the contents of the KGs were all heavily influenced by what information was most accessible---which in turn was influenced by the contents of existing structured databases.

The availability of these KGs led to NLP research in answering questions using KBs \cite{berant2013semantic,yih2016value,dubey2019lc}, or KBQA.  The most widely-used KBQA benchmarks contain only questions answerable using the KGs, and while it is difficult to estimate how many ``reasonable'' user questions are answerable with any dataset, evidence suggests it is a small minority.\footnote{The WebQuestions dataset was produced by finding questions using Google Suggest and submitting them to crowdworkers with instructions to answer them using only FreeBase, and of 100,000 submitted questions, less than 7,000 were finally included in the dataset; in another study \cite{balachandran-etal-2021-investigating}, less than 7,000 of the 307,000 training questions in the NQ dataset were included in a subset of questions likely to be answerable using WikiData.}
In contrast, \qedb{}s are designed to exploit datasets of representative user questions in determining what information will be stored in the KB, and designed to store that information in a form that facilitates matching to user questions.

Although \qedb{}s will contain different content than existing symbolic KBs, we note that they are very compatible: in fact, it is straightforward to cross-link a \qedb{} with an existing KB, by simply running an entity linker on the underlying corpus.  Such a hybrid KB would allow rather interesting combinations of reasoning---e.g., one could answer a multihop question by performing some hops in the \qedb{} and some in the symbolic KB.  In short, \qedb{}s are best thought of as a complement to existing KBs than a replacement.

\subsection{Open IE}

Open information extraction (open IE) aims to extract KBs from text using an open vocabulary of entities and relations \cite{etzioni2008open,mausam2016open}---i.e., open IE extractions are organized into a graph of document spans that refer to real world entities or describe relationships.
A \qedb{} is syntactically similar, in that it is a graph of textual spans, but the contents of the graph are not  determined by the intuitions\footnote{In fact recent open IE systems are tuned on benchmarks which specify target extractions for sample text, and these benchmarks are closely related to annotations for the semantic role labeling task \cite{bhardwaj-etal-2019-carb}.} of the system designers, as in open IE or conventional IE, but by concrete data about likely user queries.  

\subsection{RePAQ, QAMAT, and other uses of QA data} \label{sec:paq}

Our proposal is heavily influenced by PAQ \cite{lewis2021paq}.  PAQ is a resource containing 65 million question-answer pairs, and the accompanying program RePAQ is a QA system that answers a user question $q$ by finding a matching questions $q'$ in PAQ and proposing the answer $a'$ that is paired with $q'$.  Recently \citet{chen2022augmenting} described QAMAT, a similar system that outperforms RePAQ, and can also answer multi-hop questions.  As noted above we made use of the PAQ resource in producing \qedb{}.

Unlike PAQ, which is a collection of question-answer pairs, \qedb{} is structured as a KB, with free-text analogs of entities and relations. PAQ/RePAQ are aimed at single-hop QA tasks, while KBs have been used for many purposes: for example, a 2019 survey of applications of Wikidata \cite{mora2019systematic} discusses tasks like news aggregation, review annotation, evaluating trustworthiness of online information sources, entity disambiguation, bibliometrics, curation of genomic data, and studies of gender disparity over time.

More precisely, \qedb{} extends PAQ/RePAQ in several fundamental ways.  
First, unlike PAQ, it includes relational information, making it possible to answer more complex questions---for example, multi-hop questions, which require navigation through the KB. 
Second, a \qedb{} can also be easily cross-linked to an existing KB, by entity-linking the underlying corpus to the KB.

We also note that question data has been proposed as a resource for encoding generally useful language models \cite{jia2021question,he2019quase}.  While this goal is different from the one pursued here, it shares the intuition that the factual content of documents is summarized well by questions and answers.

\subsection{Phrase-indexed QA and virtual KBs}

PAQ/RePAQ are quite similar to phrase-indexed QA (PIQA) \cite{seo2018phrase,lee2020learning}.  In PIQA systems, questions are answered by matching a neural encoding of a question $q$ with a neural encoding of an answer span $a$.  This process is fast at query time, but requires one to first index all possible answer spans in a corpus.  In RePAQ, a question $q'$ is retrieved from PAQ using a similar method: one matches a neural encoding of $q$ against a previously-indexed neural encoding of $q'$.  Hence RePAQ/PAQ can be viewed as a variant of PIQA, in which rather than encoding and indexing a possible answer span $a'$ directly, one encodes and indexes a question $q'$ generated from $a'$ by a QG system.

There are advantages and disadvantages of RePAQ/PAQ's strategy of encoding an answer span indirectly via a generated question, rather than directly with neural span encoder.  The disadvantages are that  the system is also more complex, and that an additional computational cost is incurred in running QG.  The advantage is that now both steps of the encoding are semantically meaningful and verifiable: QG is correct if it produces a meaningful question that is answerable using the document, and question retrieval is correct if it finds a $q$ that is semantically equivalent to the user query $q$.  These same differences apply to \qedb{}---as well as having the fundamental differences discussed above in Section~\ref{sec:paq}, such as encoding relational information and supporting multihop reasoning.

There are also relational extensions of PIQA: for instance, DrKIT \cite{dhingra2020differentiable} indexes answer phrases (in this case, only entity mentions are considered) with additional information about their relationship to nearby entities; and OPQL \cite{sun2021reasoning} builds an index of pairs of entity mentions, rather than single mentions, again preserving information about the relationship between mentions.  Both of these systems allow multihop reasoning, and both allow an open set of relationships (but not entities).  \qedb{} is related to these systems as PAQ/RePAQ is related to PIQA: the encoding and indexing process is more expensive, because of the QG step, but also more verifiable.

\subsection{Explainable QA}

This proposal builds on work on explainable QA \cite{lamm2021qed}, and in fact relies on data collected for explainable QA projects 
to model the alignment of question references to document references.

The primary difference between a \qedb{} and explainable QA systems is that explainable QA models are applied at query time, i.e., in a ``lazy'' way, while in building a \qedb{} models are applied once to an entire corpus, i.e., in an ``eager'' way.  As a result a \qedb{} has advantages for tasks that require low latency at query time.  For example, we believe that a \qedb{} will be advantageous for questions that require ``reasoning'', in the broad sense of aggregation of information across many documents. Such questions are often expensive to answer with iterative retrieve-and-read systems.  Examples of such questions include multihop QA; finding multiple answers for an ambiguous question (e.g., ``\textit{who played jamie lannister in GoT}''); and finding many answers to a question (e.g., ``\textit{what movies did william shatner play kirk in}'').

\subsection{Automated question generation}


Many prior researchers have used QG for other applications.  QG is widely used as a way of augmented data for QA systems \cite{yang2017semi,li2020harvesting,pan2020unsupervised}---so widely used that the QG problem has received attention of the text generation community as a stand-alone task \cite{pan2020semantic}.
Although we use QG as an important part of \qedb{} construction, our end goal is different, as our goal is to extract a KB from text rather than to train a QA system.

Other researchers \cite{durmus-etal-2020-feqa,honovich2021q} have used QG to measure the factuality of generated summaries. In particular, they generate questions from a summary, and answer them using both the summary and the original source document.  If these two answers disagree, this suggests that the summary has diverged from the source in content.  Our approach also relies on QG to identify important factual statements in text, but our focus is on extracting knowledge from multiple documents, not comparing knowledge from two different documents.

\subsection{Question-based document annotation}

Prior researchers have proposed question-answer pairs as a scheme for annotating text for specific NLP tasks, e.g., for relation extraction \cite{levy2017zero},  slot filling \cite{du-etal-2021-qa} semantic role labeling  \cite{fitzgerald2018large,klein-etal-2020-qanom} and discourse relations \cite{pyatkin-etal-2020-qadiscourse}.  The advantage of this approach is that by reducing an NLP task to QA, one immediately obtains a strong baseline method for that task (since good trainable QA methods exist), and also obtains a new way to obtain annotated data, since question-answer pairs are generally easy for unskilled crowd workers to produce.  

Most related to this work, \cite{michael2017crowdsourcing} proposed question-answer pairs as a general meaning representation for text.  Specifically, this work introduced \textit{question-answer meaning representation} (QAMR) as a representation of the meaning of documents, developed a corpus of 5,000 sentences with 100,000 annotated questions, and showed that the questions covered most of the relationships in existing datasets (like PropBank, NomBank, QA-SRL, and AMR) along with many new ones.

Our work differs in emphasis from QAMR in several important ways.  First, unlike a \qedb{}, a QAMR is not intended to model what is salient in a document: rather than extracting question-answer pairs that are statistically likely given a dataset, the intent is to extract all information.  Second, QAMR was intended to model information in a single sentence, rather than in a collection of sentences, as in \qedb{}.  Because of this there was no focus on cross-document co-references (or, in fact, intra-document cross-reference), and no evaluation on tasks that require any sort of aggregation across documents: for example, while the ability to automatically answer QAMR questions was measured, this QA task was done with a known context document (i.e., a ``machine reading setting'', rather than in an ``open QA'' setting). 

QA-Align \cite{DBLP:journals/corr/abs-2109-12655} makes use of QA-SRL annotations \cite{fitzgerald2018large} as features for cross-document co-reference.  Because cross-document co-reference is one of several components in a \qedb{}, our proposal is broader in scope, but the results of \cite{DBLP:journals/corr/abs-2109-12655} support our general claim that there are synergies between the various subtasks involved in building (or using) a \qedb{}.


\section{Conclusion}

Symbolic KBs organize information into small modular components (e.g., entities, KG triples, WikiData statements) that can be combined compositionally to answer complex queries.  While many recent papers have focused on tasks like open QA, where questions are answered from text without using a KB, broad-coverage symbolic KBs continue to be widely used in practice, and, despite recent progress in methods for ``multi-hop'' QA, are still the only computationally efficient way of answering questions that combine information for multiple documents.
However, the broad-coverage KBs that are currently in wide use are largely collections of information that \emph{easily collected and integrated}, and need not reflect the actual information needs of users.
In this position paper, we advocate for a new approach to constructing KBs, and in particular, an approach to collecting modular, compositionally-combinable knowledge components from text, driven by a sample of user's questions and answers.  

Our approach begins with data on likely questions and answers, and extrapolates this data, using neural question generation, to a larger set of QA pairs.  A rich relational structure is then created by aligning the entities mentioned in these questions with traditional KB entities.  This alignment is done using data for explainable QA systems \cite{kwiatkowski2019natural} and standard entity-linking methods, and leads to a \qedb{} structure which is relational; grounded in a corpus; aligned with the original sample of questions; and easily combined with existing symbolic KBs.  

The central claims made are that like a traditional symbolic KB, a \qedb{} is composed of modular components that can be combined in compositionally in many different ways; and that the information in a \qedb{} will be more useful for answering user information needs than the information in a traditionally constructed broad-coverage KB.
The former claim is supported qualitatively, by presenting examples of the component elements of the \qedb{}, and examples of how they can be recombined.  The latter claim is supported quantatively: using available benchmarks, we show that \qedb{} is a better substrate for question-answering than symbolic KBs, unless the set of question is carefully filtered to contain only questions supported by the KB (as is the case for WebQSP).  However, even for questions chosen to be conducive to use of a KB, a \qedb{} is extremely competitive when similar QA system architectures are used: the only cases in which the traditional KB is more than 3 point better than \qedb{} is when the match between the questions and KB is ideal (WebQSP/WD) and when QA architectures are optimized for KBQA.

If these positions are generally true, this would suggest a number of new directions and emphases for research.  For example, it suggests new focuses for question generation, such as generation of high-recall sets of questions, and also suggests more work on direct evaluation of the correctness of generated questions. (Most current work evaluates QG indirectly, by its contribution to tasks like pre-training for QA, or evaluating factuality of generated text).  More notably, it suggests that the target for information extraction systems should generally be not a traditional KB, but a \qedb{}.  

\bibliography{biblio}

\begin{thebibliography}{50}
\providecommand{\natexlab}[1]{#1}
\providecommand{\url}[1]{\texttt{#1}}
\expandafter\ifx\csname urlstyle\endcsname\relax
  \providecommand{\doi}[1]{doi: #1}\else
  \providecommand{\doi}{doi: \begingroup \urlstyle{rm}\Url}\fi

\bibitem[Auer et~al.(2007)Auer, Bizer, Kobilarov, Lehmann, Cyganiak, and
  Ives]{auer2007dbpedia}
S{\"o}ren Auer, Christian Bizer, Georgi Kobilarov, Jens Lehmann, Richard
  Cyganiak, and Zachary Ives.
\newblock Dbpedia: A nucleus for a web of open data.
\newblock In \emph{The semantic web}, pages 722--735. Springer, 2007.

\bibitem[Balachandran et~al.(2021)Balachandran, Dhingra, Sun, Collins, and
  Cohen]{balachandran-etal-2021-investigating}
Vidhisha Balachandran, Bhuwan Dhingra, Haitian Sun, Michael Collins, and
  William Cohen.
\newblock Investigating the effect of background knowledge on natural
  questions.
\newblock In \emph{Proceedings of Deep Learning Inside Out (DeeLIO): The 2nd
  Workshop on Knowledge Extraction and Integration for Deep Learning
  Architectures}, pages 25--30, Online, June 2021. Association for
  Computational Linguistics.
\newblock \doi{10.18653/v1/2021.deelio-1.3}.
\newblock URL \url{https://aclanthology.org/2021.deelio-1.3}.

\bibitem[Berant et~al.(2013)Berant, Chou, Frostig, and
  Liang]{berant2013semantic}
Jonathan Berant, Andrew Chou, Roy Frostig, and Percy Liang.
\newblock Semantic parsing on freebase from question-answer pairs.
\newblock In \emph{Proceedings of the 2013 conference on empirical methods in
  natural language processing}, pages 1533--1544, 2013.

\bibitem[Bhardwaj et~al.(2019)Bhardwaj, Aggarwal, and
  Mausam]{bhardwaj-etal-2019-carb}
Sangnie Bhardwaj, Samarth Aggarwal, and Mausam Mausam.
\newblock {C}a{RB}: A crowdsourced benchmark for open {IE}.
\newblock In \emph{Proceedings of the 2019 Conference on Empirical Methods in
  Natural Language Processing and the 9th International Joint Conference on
  Natural Language Processing (EMNLP-IJCNLP)}, pages 6262--6267, Hong Kong,
  China, November 2019. Association for Computational Linguistics.
\newblock \doi{10.18653/v1/D19-1651}.
\newblock URL \url{https://aclanthology.org/D19-1651}.

\bibitem[Bollacker et~al.(2008)Bollacker, Evans, Paritosh, Sturge, and
  Taylor]{bollacker2008freebase}
Kurt Bollacker, Colin Evans, Praveen Paritosh, Tim Sturge, and Jamie Taylor.
\newblock Freebase: a collaboratively created graph database for structuring
  human knowledge.
\newblock In \emph{Proceedings of the 2008 ACM SIGMOD international conference
  on Management of data}, pages 1247--1250, 2008.

\bibitem[Chen et~al.(2022)Chen, Verga, de~Jong, Wieting, and
  Cohen]{chen2022augmenting}
Wenhu Chen, Pat Verga, Michiel de~Jong, John Wieting, and William Cohen.
\newblock Augmenting pre-trained language models with qa-memory for open-domain
  question answering.
\newblock \emph{arXiv preprint arXiv:2204.04581}, 2022.

\bibitem[Dhingra et~al.(2020)Dhingra, Zaheer, Balachandran, Neubig,
  Salakhutdinov, and Cohen]{dhingra2020differentiable}
Bhuwan Dhingra, Manzil Zaheer, Vidhisha Balachandran, Graham Neubig, Ruslan
  Salakhutdinov, and William~W Cohen.
\newblock Differentiable reasoning over a virtual knowledge base.
\newblock \emph{arXiv preprint arXiv:2002.10640}, 2020.

\bibitem[Du et~al.(2021)Du, He, Li, Yu, Pasupat, and Zhang]{du-etal-2021-qa}
Xinya Du, Luheng He, Qi~Li, Dian Yu, Panupong Pasupat, and Yuan Zhang.
\newblock {QA}-driven zero-shot slot filling with weak supervision pretraining.
\newblock In \emph{Proceedings of the 59th Annual Meeting of the Association
  for Computational Linguistics and the 11th International Joint Conference on
  Natural Language Processing (Volume 2: Short Papers)}, pages 654--664,
  Online, August 2021. Association for Computational Linguistics.
\newblock \doi{10.18653/v1/2021.acl-short.83}.
\newblock URL \url{https://aclanthology.org/2021.acl-short.83}.

\bibitem[Dubey et~al.(2019)Dubey, Banerjee, Abdelkawi, and
  Lehmann]{dubey2019lc}
Mohnish Dubey, Debayan Banerjee, Abdelrahman Abdelkawi, and Jens Lehmann.
\newblock Lc-quad 2.0: A large dataset for complex question answering over
  wikidata and dbpedia.
\newblock In \emph{International semantic web conference}, pages 69--78.
  Springer, 2019.

\bibitem[Durmus et~al.(2020)Durmus, He, and Diab]{durmus-etal-2020-feqa}
Esin Durmus, He~He, and Mona Diab.
\newblock {FEQA}: A question answering evaluation framework for faithfulness
  assessment in abstractive summarization.
\newblock In \emph{Proceedings of the 58th Annual Meeting of the Association
  for Computational Linguistics}, pages 5055--5070, Online, July 2020.
  Association for Computational Linguistics.
\newblock \doi{10.18653/v1/2020.acl-main.454}.
\newblock URL \url{https://aclanthology.org/2020.acl-main.454}.

\bibitem[Etzioni et~al.(2008)Etzioni, Banko, Soderland, and
  Weld]{etzioni2008open}
Oren Etzioni, Michele Banko, Stephen Soderland, and Daniel~S Weld.
\newblock Open information extraction from the web.
\newblock \emph{Communications of the ACM}, 51\penalty0 (12):\penalty0 68--74,
  2008.

\bibitem[F{\'e}vry et~al.(2020)F{\'e}vry, Soares, FitzGerald, Choi, and
  Kwiatkowski]{fevry2020entities}
Thibault F{\'e}vry, Livio~Baldini Soares, Nicholas FitzGerald, Eunsol Choi, and
  Tom Kwiatkowski.
\newblock Entities as experts: Sparse memory access with entity supervision.
\newblock \emph{arXiv preprint arXiv:2004.07202}, 2020.

\bibitem[FitzGerald et~al.(2018)FitzGerald, Michael, He, and
  Zettlemoyer]{fitzgerald2018large}
Nicholas FitzGerald, Julian Michael, Luheng He, and Luke Zettlemoyer.
\newblock Large-scale {QA-SRL} parsing.
\newblock \emph{arXiv preprint arXiv:1805.05377}, 2018.

\bibitem[He et~al.(2019)He, Ning, and Roth]{he2019quase}
Hangfeng He, Qiang Ning, and Dan Roth.
\newblock Quase: Question-answer driven sentence encoding.
\newblock \emph{arXiv preprint arXiv:1909.00333}, 2019.

\bibitem[Hearst et~al.(1998)Hearst, Levy, Knoblock, Minton, and
  Cohen]{hearst1998information}
Marti~A. Hearst, AY~Levy, C~Knoblock, S~Minton, and W~Cohen.
\newblock Information integration.
\newblock \emph{IEEE Intelligent Systems and their Applications}, 13\penalty0
  (5):\penalty0 12--24, 1998.

\bibitem[Honovich et~al.(2021)Honovich, Choshen, Aharoni, Neeman, Szpektor, and
  Abend]{honovich2021q}
Or~Honovich, Leshem Choshen, Roee Aharoni, Ella Neeman, Idan Szpektor, and Omri
  Abend.
\newblock {Q$^2$}: Evaluating factual consistency in knowledge-grounded
  dialogues via question generation and question answering.
\newblock \emph{arXiv preprint arXiv:2104.08202}, 2021.

\bibitem[Izacard and Grave(2020)]{izacard2020leveraging}
Gautier Izacard and Edouard Grave.
\newblock Leveraging passage retrieval with generative models for open domain
  question answering.
\newblock \emph{arXiv preprint arXiv:2007.01282}, 2020.

\bibitem[Jia et~al.(2021)Jia, Lewis, and Zettlemoyer]{jia2021question}
Robin Jia, Mike Lewis, and Luke Zettlemoyer.
\newblock Question answering infused pre-training of general-purpose
  contextualized representations.
\newblock \emph{arXiv preprint arXiv:2106.08190}, 2021.

\bibitem[Joshi et~al.(2017)Joshi, Choi, Weld, and
  Zettlemoyer]{joshi2017triviaqa}
Mandar Joshi, Eunsol Choi, Daniel~S Weld, and Luke Zettlemoyer.
\newblock Triviaqa: A large scale distantly supervised challenge dataset for
  reading comprehension.
\newblock \emph{arXiv preprint arXiv:1705.03551}, 2017.

\bibitem[Karpukhin et~al.(2020)Karpukhin, O{\u{g}}uz, Min, Lewis, Wu, Edunov,
  Chen, and Yih]{karpukhin2020dense}
Vladimir Karpukhin, Barlas O{\u{g}}uz, Sewon Min, Patrick Lewis, Ledell Wu,
  Sergey Edunov, Danqi Chen, and Wen-tau Yih.
\newblock Dense passage retrieval for open-domain question answering.
\newblock \emph{arXiv preprint arXiv:2004.04906}, 2020.

\bibitem[Klein et~al.(2020)Klein, Mamou, Pyatkin, Stepanov, He, Roth,
  Zettlemoyer, and Dagan]{klein-etal-2020-qanom}
Ayal Klein, Jonathan Mamou, Valentina Pyatkin, Daniela Stepanov, Hangfeng He,
  Dan Roth, Luke Zettlemoyer, and Ido Dagan.
\newblock {QAN}om: Question-answer driven {SRL} for nominalizations.
\newblock In \emph{Proceedings of the 28th International Conference on
  Computational Linguistics}, pages 3069--3083, Barcelona, Spain (Online),
  December 2020. International Committee on Computational Linguistics.
\newblock \doi{10.18653/v1/2020.coling-main.274}.
\newblock URL \url{https://aclanthology.org/2020.coling-main.274}.

\bibitem[Kwiatkowski et~al.(2019)Kwiatkowski, Palomaki, Redfield, Collins,
  Parikh, Alberti, Epstein, Polosukhin, Devlin, Lee,
  et~al.]{kwiatkowski2019natural}
Tom Kwiatkowski, Jennimaria Palomaki, Olivia Redfield, Michael Collins, Ankur
  Parikh, Chris Alberti, Danielle Epstein, Illia Polosukhin, Jacob Devlin,
  Kenton Lee, et~al.
\newblock Natural questions: a benchmark for question answering research.
\newblock \emph{Transactions of the Association for Computational Linguistics},
  7:\penalty0 453--466, 2019.

\bibitem[Lamm et~al.(2021)Lamm, Palomaki, Alberti, Andor, Choi, Soares, and
  Collins]{lamm2021qed}
Matthew Lamm, Jennimaria Palomaki, Chris Alberti, Daniel Andor, Eunsol Choi,
  Livio~Baldini Soares, and Michael Collins.
\newblock Qed: A framework and dataset for explanations in question answering.
\newblock \emph{Transactions of the Association for Computational Linguistics},
  9:\penalty0 790--806, 2021.

\bibitem[Lee et~al.(2020)Lee, Sung, Kang, and Chen]{lee2020learning}
Jinhyuk Lee, Mujeen Sung, Jaewoo Kang, and Danqi Chen.
\newblock Learning dense representations of phrases at scale.
\newblock \emph{arXiv preprint arXiv:2012.12624}, 2020.

\bibitem[Lee et~al.(2019)Lee, Chang, and Toutanova]{lee2019latent}
Kenton Lee, Ming-Wei Chang, and Kristina Toutanova.
\newblock Latent retrieval for weakly supervised open domain question
  answering.
\newblock \emph{arXiv preprint arXiv:1906.00300}, 2019.

\bibitem[Levy et~al.(2017)Levy, Seo, Choi, and Zettlemoyer]{levy2017zero}
Omer Levy, Minjoon Seo, Eunsol Choi, and Luke Zettlemoyer.
\newblock Zero-shot relation extraction via reading comprehension.
\newblock \emph{arXiv preprint arXiv:1706.04115}, 2017.

\bibitem[Lewis et~al.(2021)Lewis, Wu, Liu, Minervini, K{\"u}ttler, Piktus,
  Stenetorp, and Riedel]{lewis2021paq}
Patrick Lewis, Yuxiang Wu, Linqing Liu, Pasquale Minervini, Heinrich
  K{\"u}ttler, Aleksandra Piktus, Pontus Stenetorp, and Sebastian Riedel.
\newblock Paq: 65 million probably-asked questions and what you can do with
  them.
\newblock \emph{arXiv preprint arXiv:2102.07033}, 2021.

\bibitem[Li et~al.(2020)Li, Wang, Dong, Wei, and Xu]{li2020harvesting}
Zhongli Li, Wenhui Wang, Li~Dong, Furu Wei, and Ke~Xu.
\newblock Harvesting and refining question-answer pairs for unsupervised qa.
\newblock \emph{arXiv preprint arXiv:2005.02925}, 2020.

\bibitem[Mao et~al.(2020)Mao, He, Liu, Shen, Gao, Han, and
  Chen]{mao2020generation}
Yuning Mao, Pengcheng He, Xiaodong Liu, Yelong Shen, Jianfeng Gao, Jiawei Han,
  and Weizhu Chen.
\newblock Generation-augmented retrieval for open-domain question answering.
\newblock \emph{arXiv preprint arXiv:2009.08553}, 2020.

\bibitem[Mausam(2016)]{mausam2016open}
Mausam Mausam.
\newblock Open information extraction systems and downstream applications.
\newblock In \emph{Proceedings of the twenty-fifth international joint
  conference on artificial intelligence}, pages 4074--4077, 2016.

\bibitem[Michael et~al.(2017)Michael, Stanovsky, He, Dagan, and
  Zettlemoyer]{michael2017crowdsourcing}
Julian Michael, Gabriel Stanovsky, Luheng He, Ido Dagan, and Luke Zettlemoyer.
\newblock Crowdsourcing question-answer meaning representations.
\newblock \emph{arXiv preprint arXiv:1711.05885}, 2017.

\bibitem[Mitchell et~al.(2018)Mitchell, Cohen, Hruschka, Talukdar, Yang,
  Betteridge, Carlson, Dalvi, Gardner, Kisiel, et~al.]{mitchell2018never}
Tom Mitchell, William Cohen, Estevam Hruschka, Partha Talukdar, Bishan Yang,
  Justin Betteridge, Andrew Carlson, Bhavana Dalvi, Matt Gardner, Bryan Kisiel,
  et~al.
\newblock Never-ending learning.
\newblock \emph{Communications of the ACM}, 61\penalty0 (5):\penalty0 103--115,
  2018.

\bibitem[Mora-Cantallops et~al.(2019)Mora-Cantallops, S{\'a}nchez-Alonso, and
  Garc{\'\i}a-Barriocanal]{mora2019systematic}
Mar{\c{c}}al Mora-Cantallops, Salvador S{\'a}nchez-Alonso, and Elena
  Garc{\'\i}a-Barriocanal.
\newblock A systematic literature review on wikidata.
\newblock \emph{Data Technologies and Applications}, 53\penalty0 (3):\penalty0
  250--268, 2019.

\bibitem[Pan et~al.(2020{\natexlab{a}})Pan, Chen, Xiong, Kan, and
  Wang]{pan2020unsupervised}
Liangming Pan, Wenhu Chen, Wenhan Xiong, Min-Yen Kan, and William~Yang Wang.
\newblock Unsupervised multi-hop question answering by question generation.
\newblock \emph{arXiv preprint arXiv:2010.12623}, 2020{\natexlab{a}}.

\bibitem[Pan et~al.(2020{\natexlab{b}})Pan, Xie, Feng, Chua, and
  Kan]{pan2020semantic}
Liangming Pan, Yuxi Xie, Yansong Feng, Tat-Seng Chua, and Min-Yen Kan.
\newblock Semantic graphs for generating deep questions.
\newblock \emph{arXiv preprint arXiv:2004.12704}, 2020{\natexlab{b}}.

\bibitem[Pyatkin et~al.(2020)Pyatkin, Klein, Tsarfaty, and
  Dagan]{pyatkin-etal-2020-qadiscourse}
Valentina Pyatkin, Ayal Klein, Reut Tsarfaty, and Ido Dagan.
\newblock {QAD}iscourse - {D}iscourse {R}elations as {QA} {P}airs:
  {R}epresentation, {C}rowdsourcing and {B}aselines.
\newblock In \emph{Proceedings of the 2020 Conference on Empirical Methods in
  Natural Language Processing (EMNLP)}, pages 2804--2819, Online, November
  2020. Association for Computational Linguistics.
\newblock \doi{10.18653/v1/2020.emnlp-main.224}.
\newblock URL \url{https://aclanthology.org/2020.emnlp-main.224}.

\bibitem[Rebele et~al.(2016)Rebele, Suchanek, Hoffart, Biega, Kuzey, and
  Weikum]{rebele2016yago}
Thomas Rebele, Fabian Suchanek, Johannes Hoffart, Joanna Biega, Erdal Kuzey,
  and Gerhard Weikum.
\newblock Yago: A multilingual knowledge base from wikipedia, wordnet, and
  geonames.
\newblock In \emph{International semantic web conference}, pages 177--185.
  Springer, 2016.

\bibitem[Roth et~al.(2014)Roth, Ji, Chang, and Cassidy]{roth2014wikification}
Dan Roth, Heng Ji, Ming-Wei Chang, and Taylor Cassidy.
\newblock Wikification and beyond: The challenges of entity and concept
  grounding.
\newblock \emph{ACL (Tutorial Abstracts)}, 7, 2014.

\bibitem[Sarawagi(2008)]{sarawagi2008information}
Sunita Sarawagi.
\newblock \emph{Information extraction}.
\newblock Now Publishers Inc, 2008.

\bibitem[Seo et~al.(2018)Seo, Kwiatkowski, Parikh, Farhadi, and
  Hajishirzi]{seo2018phrase}
Minjoon Seo, Tom Kwiatkowski, Ankur~P Parikh, Ali Farhadi, and Hannaneh
  Hajishirzi.
\newblock Phrase-indexed question answering: A new challenge for scalable
  document comprehension.
\newblock \emph{arXiv preprint arXiv:1804.07726}, 2018.

\bibitem[Sun et~al.(2021)Sun, Verga, Dhingra, Salakhutdinov, and
  Cohen]{sun2021reasoning}
Haitian Sun, Pat Verga, Bhuwan Dhingra, Ruslan Salakhutdinov, and William~W
  Cohen.
\newblock Reasoning over virtual knowledge bases with open predicate relations.
\newblock \emph{arXiv preprint arXiv:2102.07043}, 2021.

\bibitem[Verga et~al.(2021)Verga, Sun, Soares, and Cohen]{verga2021adaptable}
Pat Verga, Haitian Sun, Livio~Baldini Soares, and William Cohen.
\newblock Adaptable and interpretable neural memoryover symbolic knowledge.
\newblock In \emph{Proceedings of the 2021 Conference of the North American
  Chapter of the Association for Computational Linguistics: Human Language
  Technologies}, pages 3678--3691, 2021.

\bibitem[Vrande{\v{c}}i{\'c} and Kr{\"o}tzsch(2014)]{vrandevcic2014wikidata}
Denny Vrande{\v{c}}i{\'c} and Markus Kr{\"o}tzsch.
\newblock Wikidata: a free collaborative knowledgebase.
\newblock \emph{Communications of the ACM}, 57\penalty0 (10):\penalty0 78--85,
  2014.

\bibitem[Wang and Cohen(2008)]{wang2008iterative}
Richard~C Wang and William~W Cohen.
\newblock Iterative set expansion of named entities using the web.
\newblock In \emph{2008 eighth IEEE international conference on data mining},
  pages 1091--1096. IEEE, 2008.

\bibitem[Weiss et~al.(2021)Weiss, Roit, Klein, Ernst, and
  Dagan]{DBLP:journals/corr/abs-2109-12655}
Daniela~Brook Weiss, Paul Roit, Ayal Klein, Ori Ernst, and Ido Dagan.
\newblock Qa-align: Representing cross-text content overlap by aligning
  question-answer propositions.
\newblock \emph{CoRR}, abs/2109.12655, 2021.
\newblock URL \url{https://arxiv.org/abs/2109.12655}.

\bibitem[Wieting et~al.(2019)Wieting, Neubig, and
  Berg-Kirkpatrick]{wieting2019bilingual}
John Wieting, Graham Neubig, and Taylor Berg-Kirkpatrick.
\newblock A bilingual generative transformer for semantic sentence embedding.
\newblock \emph{arXiv preprint arXiv:1911.03895}, 2019.

\bibitem[Wolfson et~al.(2020)Wolfson, Geva, Gupta, Gardner, Goldberg, Deutch,
  and Berant]{wolfson2020break}
Tomer Wolfson, Mor Geva, Ankit Gupta, Matt Gardner, Yoav Goldberg, Daniel
  Deutch, and Jonathan Berant.
\newblock Break it down: A question understanding benchmark.
\newblock \emph{Transactions of the Association for Computational Linguistics},
  8:\penalty0 183--198, 2020.

\bibitem[Yang et~al.(2017)Yang, Hu, Salakhutdinov, and Cohen]{yang2017semi}
Zhilin Yang, Junjie Hu, Ruslan Salakhutdinov, and William~W Cohen.
\newblock Semi-supervised qa with generative domain-adaptive nets.
\newblock \emph{arXiv preprint arXiv:1702.02206}, 2017.

\bibitem[Yang et~al.(2018)Yang, Qi, Zhang, Bengio, Cohen, Salakhutdinov, and
  Manning]{yang2018hotpotqa}
Zhilin Yang, Peng Qi, Saizheng Zhang, Yoshua Bengio, William~W Cohen, Ruslan
  Salakhutdinov, and Christopher~D Manning.
\newblock Hotpotqa: A dataset for diverse, explainable multi-hop question
  answering.
\newblock \emph{arXiv preprint arXiv:1809.09600}, 2018.

\bibitem[Yih et~al.(2016)Yih, Richardson, Meek, Chang, and Suh]{yih2016value}
Wen-tau Yih, Matthew Richardson, Christopher Meek, Ming-Wei Chang, and Jina
  Suh.
\newblock The value of semantic parse labeling for knowledge base question
  answering.
\newblock In \emph{Proceedings of the 54th Annual Meeting of the Association
  for Computational Linguistics (Volume 2: Short Papers)}, pages 201--206,
  2016.

\end{thebibliography}
\bibliographystyle{plainnat}




\end{document}